\documentclass[runningheads]{llncs}

\usepackage[T1]{fontenc}
\usepackage{graphicx,verbatim}
\usepackage{amsmath}
\usepackage{amssymb}
\usepackage{booktabs}
\usepackage{color}
\usepackage{hyperref}

\newcommand{\Hop}{\ensuremath{\mathcal{H}_{\mathrm{op}}}}

\begin{document}

\title{When Measurement Conventions Masquerade as Calibration Gains in Cardiac Digital Twins}

\titlerunning{Measurement Conventions and EF Calibration}

\author{Dang P. M. Cao\inst{1,2}\thanks{Cao Pham Minh Dang is the corresponding author. Email: \texttt{24dang.cpm@vinuni.edu.vn}} \and
Hieu Pham\inst{1,2,3}}

\authorrunning{D. P. M. Cao and H. Pham}

\institute{
\textsuperscript{1} College of Engineering and Computer Science, VinUniversity,
Hanoi, Vietnam\\
\textsuperscript{2} VinUni-Illinois Smart Health Center, VinUniversity, Hanoi, Vietnam\\
\textsuperscript{3} Center for Innovations in Health Sciences, VinUniversity, Hanoi, Vietnam\\
\email{24dang.cpm@vinuni.edu.vn, hieu.ph@vinuni.edu.vn}
}

\maketitle

\begin{abstract}
Cardiac digital twins convert clinical images into physiological measurements through observation operators, yet calibration studies often assume a fixed reference convention. Across four shared-backbone echocardiographic EF front-ends, phase conditioning appears to remove CAMUS baseline bias. Matched-reference analysis rejects this gain: single-plane ground-truth EF error is statistically indistinguishable across models, while single-plane ground-truth EF exceeds CAMUS biplane clinical EF by $+6.30$ points, explaining nearly all baseline bias. A prespecified EchoNet-Dynamic replication, with released data and our extractor aligned to the apical four-chamber plane, removes baseline overestimation and reverses the CAMUS ranking. We also quantify haemodynamic effects, conformal residual-width budgets, and EF-stratum changes, yielding a Convention-Aware EF Audit protocol that separates genuine observation operator calibration from measurement artefacts. GitHub: \href{https://github.com/minhdang050806/Ejection-Fraction-Bias-in-Cardiac-Digital-Twin.git}{Ejection-Fraction-Bias-in-Cardiac-Digital-Twin.git}

\end{abstract}

\section{Introduction}

Cardiac digital twins integrate patient-specific data with mechanistic and statistical models~\cite{corral2020digital,niederer2021scaling,peirlinck2021precision}; recent pipelines use imaging and electrocardiography to personalise anatomy and physiology~\cite{camps2024harnessing,qian2025developing}. We call this image-to-measurement mapping the \emph{observation operator} $\Hop$. Though often treated as preprocessing, systematic error in $\Hop$ biases downstream state updates regardless of model quality.

For LV function, $\Hop$ typically segments echocardiograms before extracting volumes and EF. CAMUS~\cite{leclerc2019camus} and EchoNet-Dynamic~\cite{ouyang2020echonet} advanced LV segmentation, commonly evaluated by Dice~\cite{leclerc2019camus,ouyang2020echonet,wei2020clas}. However, Dice is geometric; high overlap can still bias EDV, ESV, or their ratio and thereby the physiological innovation supplied to a twin.

We audit echo-based EF calibration using four front-ends sharing a U-Net~\cite{ronneberger2015unet}: frame-wise B1; adjacent-frame-smoothed B2; P1, combining FiLM~\cite{perez2018film} with a phase head motivated by echocardiographic phase characterisation~\cite{dezaki2017cardiac}; and P2, adding cross-patient cyclic consistency and Mean-Teacher-style EMA averaging~\cite{tarvainen2017mean}. Rather than proposing an architecture or closed-loop twin, we isolate whether front-end design genuinely improves EF calibration.

Reference convention is the key challenge. CAMUS reports biplane EF from apical four- and two-chamber views~\cite{leclerc2019camus}, but our extractor uses one four-chamber plane. EchoNet-Dynamic provides four-chamber videos, clinical EF/volume labels, and expert ED/ES LV tracings~\cite{ouyang2020echonet}. CAMUS is therefore plane-mismatched; EchoNet is plane-aligned but may still differ in ED/ES selection, contours, preprocessing, and calculation. A genuine gain should survive removal of the dominant plane mismatch.

We contribute: (1) an EF observation-operator audit; (2) evidence that CAMUS phase gains reflect a measured single-plane--biplane gap; (3) a negative plane-aligned EchoNet replication; and (4) quantified twin-state, residual-width, and EF-stratum consequences within a Convention-Aware EF Audit protocol.

\section{Methodology}\label{sec:method}

\subsection{EF as an observation operator}\label{sec:ef-obs}

The operator maps an apical four-chamber clip to EF. Each phase volume uses a single-plane method-of-disks approximation adapted from standard chamber quantification~\cite{lang2015recommendations}:
\begin{equation}
V=\sum_{i=1}^{N}\frac{\pi}{4}d_i^{2}\frac{L}{N},\qquad
\widehat{\mathrm{EF}}=
\frac{\widehat{\mathrm{EDV}}-\widehat{\mathrm{ESV}}}
{\widehat{\mathrm{EDV}}}\times100,
\label{eq:simpson}
\end{equation}
where $L$ is LV long-axis length and $d_i$ the $i$-th disk diameter; both datasets use this extractor.

In a Kalman formulation~\cite{kalman1960new}, EF is measurement $y$ and $\hat{y}=\Hop(\mathbf{x}^{-})$ its prediction from prior state $\mathbf{x}^{-}$. Bias $\tilde{y}=y+b$ shifts the innovation by $b$ and therefore the posterior; a zero-mean observation model cannot represent structured convention mismatch.

CAMUS provides biplane clinical EF~\cite{leclerc2019camus}, whereas EchoNet provides four-chamber videos, clinical EF/volume labels, and expert ED/ES tracings~\cite{ouyang2020echonet}. Our extractor is plane-mismatched on CAMUS and plane-aligned, but not process-matched, on EchoNet. Pixel spacing cancels in the EF ratio; its absence in EchoNet affects neither EF nor this audit, but prevents patient-specific absolute-volume propagation.

\subsection{Front-ends and metrics}

All front-ends share the U-Net backbone, optimiser, augmentation, and early stopping. \textbf{B1} is frame-wise; \textbf{B2} adds symmetric-KL adjacent-frame smoothing; \textbf{P1} combines FiLM~\cite{perez2018film} with a normalised phase head~\cite{dezaki2017cardiac}; and \textbf{P2} adds phase-matched cross-patient cyclic consistency and Mean-Teacher-style EMA~\cite{tarvainen2017mean}. AdamW~\cite{loshchilov2019decoupled} used $\mathrm{lr}=10^{-4}$, batch size 4, and 600/400 CAMUS/EchoNet epochs. P1/P2 require two non-streaming passes; B2 streams.

We report \textbf{EF MAE$_{\mathrm{gt}}$} against ground-truth-mask EF from the same extractor, isolating segmentation error; \textbf{EF MAE$_{\mathrm{clin}}$} against clinical EF, including convention mismatch; and \textbf{EF MAE$_{\mathrm{auto}}$} using ED/ES frames selected by maximum/minimum predicted LV volume.

Before EchoNet aggregation, we specified four checks: a genuine CAMUS effect should preserve baseline positive bias, phase-related bias reduction, comparable segmentation, and extractor-matched EF advantage.

\section{Experiments and Results}

\subsection{Evaluation protocol}

For CAMUS~\cite{leclerc2019camus}, we evaluate on our held-out validation subset of $n=50$ patients drawn from the 450 released training cases; clinical EF uses the apical four- and two-chamber views. For EchoNet-Dynamic~\cite{ouyang2020echonet}, we use the official test split of 1,288 studies and exclude one study with internally inconsistent clinical volume labels ($\mathrm{ESV}>\mathrm{EDV}$), yielding $n=1{,}287$. This exclusion was specified before model evaluation. All clips are resampled to 32 frames at $256\times256$ and $z$-score normalised. We use paired Wilcoxon tests, bootstrap 95\% confidence intervals, Cohen's~$d_z$, and Holm correction.

\subsection{CAMUS: apparent calibration gain with two warning signs}\label{sec:camus}

CAMUS initially suggests that phase conditioning improves EF calibration: despite similar LV Dice, B1/B2 overestimate clinical EF, whereas P1/P2 move bias toward zero (Table~\ref{tab:camus}). Evaluated only against clinical EF, this looks like a model-level gain. Two checks reject that interpretation.

\noindent\textbf{Matched-reference EF error does not improve.}
When EF is computed from the same single-plane ground-truth masks used by our extractor, EF MAE$_{\mathrm{gt}}$ is statistically indistinguishable across front-ends. The ranking also changes: P1, competitive against clinical EF, has the largest extractor-matched error and most negative bias.

\noindent\textbf{The apparent gain is seed-unstable and Dice-insensitive.}
Baselines are seed-consistent, whereas phase-conditioned models vary more. P1 shows that similar Dice need not imply calibrated EF: small phase-dependent contour shifts at ED/ES can coherently alter volumes and be amplified by the EF ratio. This is a plausible mechanism, not component attribution; FiLM, cyclic consistency, and EMA remain entangled.

\subsection{The CAMUS effect is explained by a convention offset}\label{sec:offset}

We measured the convention gap by applying our single-plane EF extractor to CAMUS ground-truth masks and comparing the resulting EF with biplane clinical EF.\footnote{One CAMUS patient was excluded because the ground-truth mask produced zero end-systolic volume; this exclusion was specified before estimating the convention offset.} Table~\ref{tab:offset} confirms the confound: the measured single-plane--biplane gap closely matches the baseline clinical bias. After subtracting this convention component, B1's residual segmentation bias has a confidence interval including zero. Thus the baseline is not substantially miscalibrated once both measurements use the same convention.

\begin{table}[!htbp]
\centering
\caption{\textbf{CAMUS} ($n=50$). EF metrics are three-seed mean$\pm$std; Dice is the seed mean. Phase gain appears only against biplane clinical EF; matched single-plane GT errors are flat.}
\label{tab:camus}
\setlength{\tabcolsep}{2.0pt}
\fontsize{8}{9}\selectfont
\begin{tabular}{lcccccc}
\hline
Model & Dice & MAE$_{\mathrm{gt}}$ & Bias$_{\mathrm{gt}}$ & MAE$_{\mathrm{clin}}$ & MAE$_{\mathrm{auto}}$ & Bias$_{\mathrm{clin}}$ \\
\hline
B1 & 0.936 & $8.47{\pm}0.81$ & $-1.37{\pm}0.58$ & $11.39{\pm}0.28$ & $13.15{\pm}0.62$ & $+4.83{\pm}0.53$ \\
B2 & 0.934 & $\mathbf{8.08{\pm}0.93}$ & $-1.91{\pm}1.09$ & $10.56{\pm}0.91$ & $11.05{\pm}1.25$ & $+4.61{\pm}1.61$ \\
P1 & 0.925 & $10.62{\pm}0.48$ & $-7.56{\pm}1.82$ & $9.97{\pm}0.35$ & $9.68{\pm}0.86$ & $-0.56{\pm}2.26$ \\
P2 & 0.932 & $8.94{\pm}1.26$ & $-6.27{\pm}0.81$ & $\mathbf{9.24{\pm}1.41}$ & $\mathbf{9.16{\pm}0.23}$ & $\mathbf{+0.37{\pm}1.31}$ \\
\hline
\end{tabular}
\end{table}

\begin{figure}[t]
\centering
\includegraphics[width=0.92\textwidth]{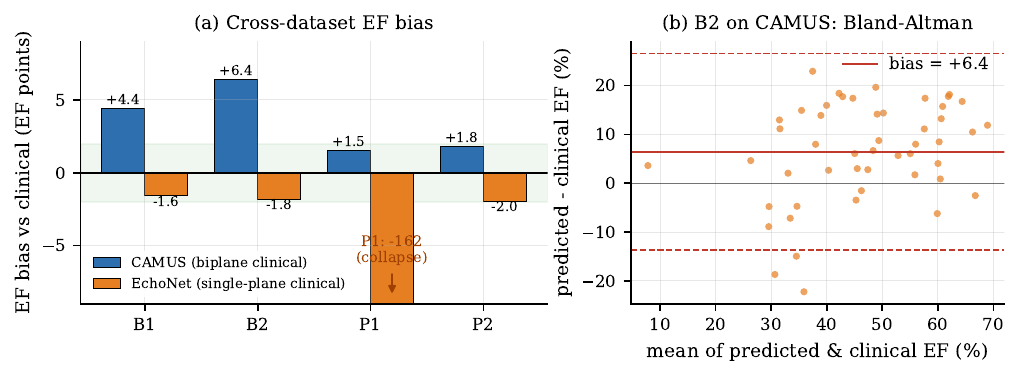}\\[2pt]
\includegraphics[width=0.50\textwidth]{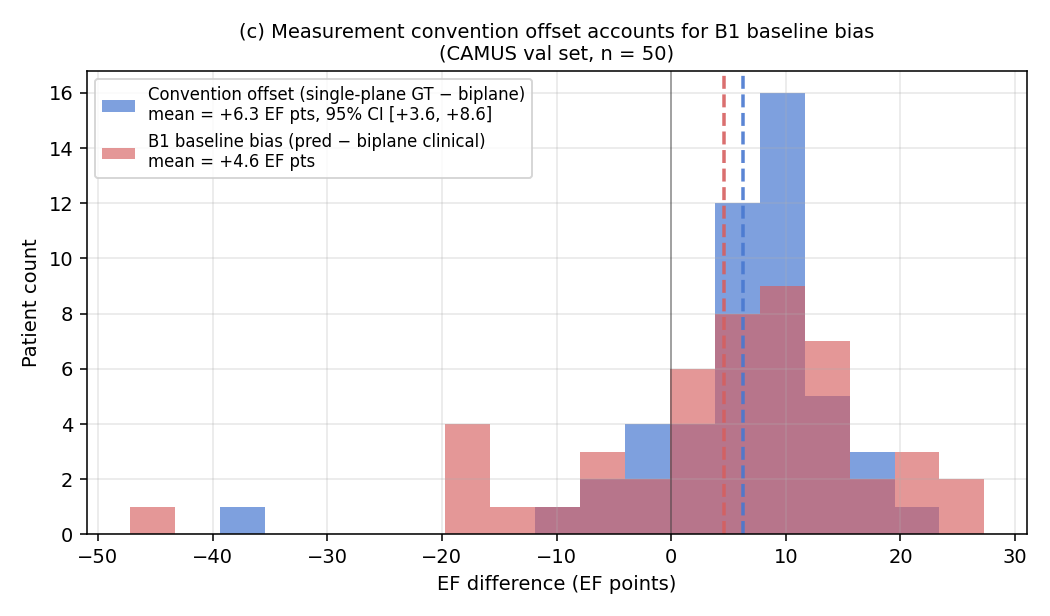}\hfill
\includegraphics[width=0.42\textwidth]{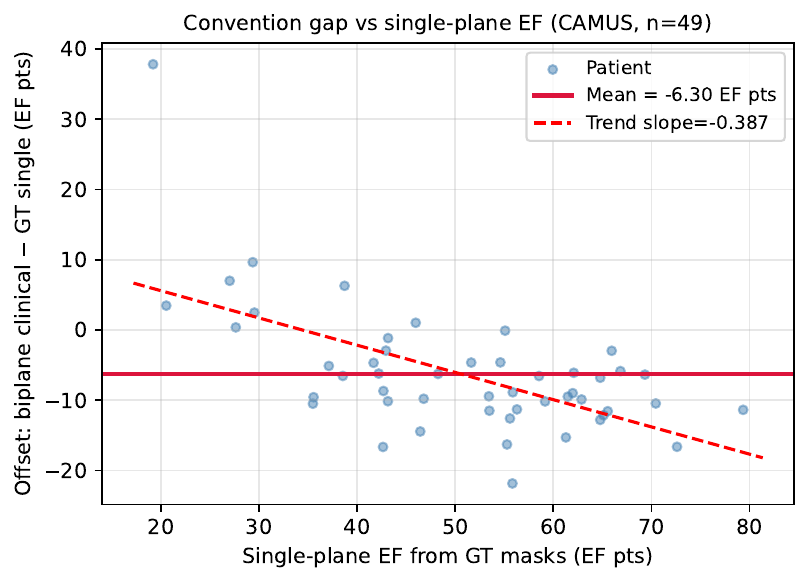}
\caption{Cross-dataset EF calibration audit. (a) CAMUS baselines overestimate clinical EF while phase-conditioned models appear near zero; plane-aligned EchoNet evaluation removes this pattern. Shaded band: ${\pm}2$ EF pts. (b) B2 Bland--Altman on CAMUS~\cite{bland1986statistical}. (c) CAMUS convention offset ($n=49$): single-plane GT EF exceeds biplane clinical EF by $+6.30$ points. (d) OLS mapping; slope $<1$ indicates a proportional gap.}
\label{fig:cross}
\end{figure}

The gap is not constant. Figure~\ref{fig:cross}(d) shows a slope below one between single-plane and biplane EF, indicating a proportional discrepancy. A constant correction removes mean bias but leaves structured residual error, especially in high-EF patients.

\begin{table}[!htbp]
\centering
\caption{\textbf{CAMUS convention-offset analysis} ($n=49$). The measured single-plane--biplane gap explains nearly all B1 clinical bias. $^\dagger$Post-hoc OLS; not a validated correction.}
\label{tab:offset}
\setlength{\tabcolsep}{4pt}
\fontsize{8}{9}\selectfont
\begin{tabular}{lcc}
\toprule
Quantity & Estimate & Uncertainty \\
\midrule
Single-plane GT $-$ biplane clinical & $+6.30$ EF pts & 95\% CI $[+3.69,+8.91]$ \\
Observed B1 clinical EF bias & $+4.83\pm0.53$ EF pts & seed std \\
Convention component & $+6.30$ EF pts & 95\% CI $[+3.69,+8.91]$ \\
Residual segmentation component & $-1.47$ EF pts & 95\% CI $[-4.08,+1.14]$ \\
Post-hoc OLS slope$^\dagger$ & 0.613 & 95\% CI $[0.47,0.76]$ \\
Post-hoc OLS intercept$^\dagger$ & $+13.34$ EF pts & -- \\
\bottomrule
\end{tabular}
\end{table}

\subsection{Plane-aligned replication on EchoNet-Dynamic}\label{sec:audit}

EchoNet-Dynamic tests the convention-confound hypothesis because labels and extractor are single-plane aligned, although their full measurement processes need not be identical. If the CAMUS result were a genuine model effect, baselines should remain positively biased and phase-conditioned models should reduce that bias. Table~\ref{tab:echonet} shows the opposite: baseline bias changes sign, the CAMUS ranking does not replicate, and B2 is numerically strongest. This reversal is expected when the dominant plane mismatch is removed.

\noindent\textbf{Deployment risk of phase-conditioned inference.}
Segmentation remains comparable, but P1/P2 produce invalid EF when $\widehat{\mathrm{ESV}}\ge\widehat{\mathrm{EDV}}$, consistent with ED/ES sensitivity in two-pass conditioning. This implementation-level result does not show that phase conditioning is generally harmful. A safety gate can flag failures; B2 remains the strongest audited deployable option, with streaming inference, no invalid EF, and the best plane-aligned EchoNet EF MAE$_{\mathrm{clin}}$.

\subsection{Downstream implications for cardiac digital twins}\label{sec:support}

\begin{figure}[t]
\centering
\includegraphics[width=0.8\textwidth]{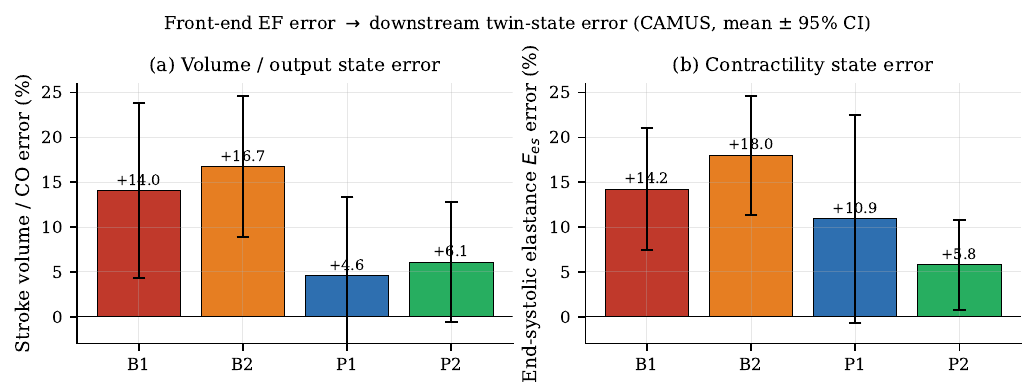}
\caption{Downstream twin-state errors under CAMUS biplane-reference conditions (mean$\pm$95\% CI), using Section~\ref{sec:support}. Under plane-aligned EchoNet conditions, SV and CO errors fall to about 3--4\%, so much of the penalty is recoverable after removing the convention gap.}
\label{fig:twin}
\end{figure}

\noindent\textbf{Haemodynamic propagation.}
We estimate downstream EF-bias effects using a simplified single-ventricle lumped calculation. With fixed reference EDV, $\Delta \mathrm{SV}=(\Delta\mathrm{EF}/100)\,\mathrm{EDV}_{\mathrm{ref}}$ and $\Delta \mathrm{CO}=\Delta \mathrm{SV}\,\mathrm{HR}_{\mathrm{ref}}/1000$, with $\mathrm{HR}_{\mathrm{ref}}=72$ bpm. End-systolic elastance error uses a simplified pressure--volume relationship motivated by Suga and Sagawa~\cite{suga1974instantaneous}. This is order-of-magnitude propagation, not a patient-specific closed-loop simulation; Figure~\ref{fig:twin} summarises the errors.

\noindent\textbf{Conformal observation-error budgets.}
We express EF residuals as split-conformal prediction intervals~\cite{vovk2005algorithmic,angelopoulos2023conformal}. We interpret their empirical half-widths as observation-error budgets for comparing the evaluated measurement pipelines, not as direct estimates of a Kalman observation covariance. Under convention-confounded evaluation, the calibration residuals contain segmentation error and structured reference mismatch; under plane-aligned evaluation, they contain segmentation error, frame-selection error, and remaining measurement-process differences. The intervals aggregate these sources rather than identifying them separately (Table~\ref{tab:conformal}).

\begin{table}[!htbp]
\centering
\caption{\textbf{EchoNet-Dynamic} ($n=1{,}287$ held-out after one prespecified exclusion). Under plane-aligned evaluation, the CAMUS phase-conditioning advantage disappears. Continuous metrics are mean$\pm$std over three seeds. Invalid EF counts $\widehat{\mathrm{ESV}}\ge\widehat{\mathrm{EDV}}$ as a separate deployment failure.}
\label{tab:echonet}
\setlength{\tabcolsep}{2.5pt}
\fontsize{8}{9}\selectfont
\begin{tabular}{lcccccc}
\toprule
Model & Seeds & Invalid EF & Dice & MAE$_{\mathrm{gt}}$ & MAE$_{\mathrm{clin}}$ & Bias$_{\mathrm{clin}}$ \\
\midrule
B1 & 3 & 0 & $\mathbf{0.925\pm0.000}$ & $6.37\pm0.11$ & $6.02\pm0.14$ & $-1.72\pm0.47$ \\
B2 & 3 & 0 & $0.924\pm0.000$ & $\mathbf{6.29\pm0.09}$ & $\mathbf{5.92\pm0.04}$ & $\mathbf{-1.41\pm0.12}$ \\
P1 & 3 & $10$--$20$ & $0.918\pm0.004$ & $8.54\pm0.50$ & $8.55\pm0.70$ & $-5.97\pm1.00$ \\
P2 & 3 & $8$--$11$ & $0.922\pm0.001$ & $6.61\pm0.21$ & $6.33\pm0.28$ & $-2.72\pm0.86$ \\
\bottomrule
\end{tabular}
\end{table}

\begin{table}[!htbp]
\centering
\caption{\textbf{EchoNet replication decision specified before aggregation.} Plane-aligned replication fails the checks expected under a genuine phase-conditioning calibration effect.}
\label{tab:replication}
\setlength{\tabcolsep}{3pt}
\fontsize{8}{9}\selectfont
\begin{tabular}{p{0.29\linewidth}p{0.27\linewidth}p{0.27\linewidth}c}
\toprule
Criterion & Expected if CAMUS gain is genuine & Observed on EchoNet & Verdict \\
\midrule
Baseline positive bias persists &
B1/B2 remain positively biased &
B1/B2 biases are negative &
Fail \\
Phase models reduce baseline bias &
P1/P2 move bias closer to zero &
P1/P2 do not improve bias over B2 &
Fail \\
Segmentation remains comparable &
Dice does not collapse &
Dice remains comparable across front-ends &
Pass \\
Extractor-matched EF advantage persists &
Phase models improve MAE$_{\mathrm{gt}}$ &
B2 has lowest MAE$_{\mathrm{gt}}$ &
Fail \\
\bottomrule
\end{tabular}
\end{table}

\noindent\textbf{EF-stratum assignment.}
Using the LVEF cut points employed in the 2022 AHA/ACC/HFSA classification---$\leq40\%$, $41$--$49\%$, and $\geq50\%$~\cite{heidenreich2022aha}---we measure disagreement between EF strata. These strata are used only for sensitivity analysis and must not be interpreted as heart-failure diagnoses. Uncorrected CAMUS convention mismatch causes frequent EF-stratum disagreement (Table~\ref{tab:staging}). P2 has the best apparent CAMUS calibration, but its disagreement rate overlaps B1 within uncertainty, so the apparent gain is not deployment-grade evidence.

\section{Convention-Aware EF Audit Protocol}\label{sec:protocol}

The results motivate a five-step Convention-Aware EF Audit (CAEA) protocol for EF observation operators. The purpose is simple: prevent a model from receiving credit for calibration gains caused by the measurement-reference pair.

\begin{enumerate}
\item \textbf{Compute EF MAE$_{\mathrm{gt}}$ under the extractor convention.}
Apply the deployed EF extractor to ground-truth masks. If MAE$_{\mathrm{gt}}$ is flat while MAE$_{\mathrm{clin}}$ varies, treat the apparent clinical-reference gain as suspicious.

\item \textbf{Characterise the convention offset.}
Estimate $\widehat{\delta}$ between GT EF under the extractor convention and clinical EF. Offsets comparable to claimed model gains are practically relevant. Check residual trends because a constant correction may remove mean bias but leave structured error.

\item \textbf{Check seed consistency.}
Run independent seeds when feasible. A clinical EF advantage with large cross-seed variation is unstable, especially for phase-conditioned models where phase-estimation shifts can change ED/ES volumes.

\item \textbf{Add a deployment safety gate.}
Flag $\widehat{\mathrm{ESV}}\geq\widehat{\mathrm{EDV}}$ as an internally inconsistent estimate. Separately flag $\widehat{\mathrm{EF}}<5\%$ as a study-specific extreme-value quality-control threshold rather than a physiologically impossible value. Both conditions should trigger review before the estimate enters a downstream twin.

\item \textbf{Replicate under an aligned reference.}
Treat gains as genuine only if they persist after aligning the measurement plane and auditing residual differences in frame selection, contours, preprocessing, and calculation.
\end{enumerate}

CAEA is architecture-independent in design; validation on direct EF regressors, video models, stronger backbones, and hybrid pipelines remains required.

\section{Discussion}\label{sec:discuss}

\noindent\textbf{Main lesson.}
Phase conditioning appeared to improve CAMUS EF calibration despite similar Dice, but the gain disappeared after plane alignment. Because EF and ventricular-volume estimates depend on imaging and measurement conventions~\cite{wood2014ejection}, calibration should be assessed for the full model--extractor--reference pair, using extractor-matched EF, convention auditing, seed consistency, and external replication.

\noindent\textbf{Clinical and digital-twin implications.}
Differences between single-plane and recommended biplane quantification are well recognised~\cite{lang2015recommendations,wood2014ejection}; here, they create a benchmark artefact that favours phase-conditioned models without improving extractor-matched EF. The audit separates convention bias from model error and quantifies downstream effects. Among the evaluated front-ends, B2 is most deployable: it supports streaming, produces no invalid EF estimates, and achieves the best plane-aligned EchoNet EF MAE$_{\mathrm{clin}}$.

\noindent\textbf{Limitations.}
The audit covers two datasets, one EF extractor, and four shared-backbone U-Net front-ends; broader datasets and direct or video-based regressors remain untested. CAMUS includes only 49--50 patients, so EF-stratum and conformal analyses are exploratory. EchoNet aligns the imaging plane but not necessarily ED/ES selection, contours, preprocessing, or calculation, and CMR validation is still needed. P2 entangles FiLM, cyclic consistency, and EMA, preventing general conclusions about phase conditioning. Haemodynamic propagation assumes fixed volumes and heart rate rather than a patient-specific closed-loop twin.

\begin{credits}
\subsubsection{\ackname}
The authors thank Mr. Ha-Hieu Pham for his valuable advice on manuscript
preparation and the refinement of the research ideas. The
computational resources used for model training were provided by the VinUni-Illinois Smart Health Center (VISHC)
Server infrastructure. This research was funded by the National Foundation for Science and Technology Development (NAFOSTED) through Project No. IZVSZ2\_229539 (2025–2027)
\end{credits}

\begin{table}[!htbp]
\centering
\caption{\textbf{Split-conformal EF observation-error budgets.} Half-width in EF points. The squared-width index is $W_{\mathrm{rel}}=(h/h_{\mathrm{EchoNet,B2}})^2$; it is a comparative scale index, not a direct estimate of Kalman observation covariance. CAMUS uses a 50/50 calibration/evaluation split of our 50-patient validation subset, so its intervals are exploratory.}
\label{tab:conformal}
\setlength{\tabcolsep}{5pt}
\fontsize{8}{9}\selectfont
\begin{tabular}{llccc}
\toprule
Dataset / reference & Model & 90\% half-width & Coverage & $W_{\mathrm{rel}}$ \\
\midrule
CAMUS clinical, biplane-mismatched & B1 & $23.46$ & $96.0\%$ & $2.89$ \\
CAMUS clinical, biplane-mismatched & P2 & $16.96$ & $100.0\%$ & $1.51$ \\
EchoNet clinical, plane-aligned & B2 & $13.80$ & $92.4\%$ & 1.00 \\
\bottomrule
\end{tabular}
\end{table}

\begin{table}[!htbp]
\centering
\caption{\textbf{EF-stratum disagreement under CAMUS convention mismatch} ($n=50$). Strata use guideline LVEF cut points for sensitivity analysis and must not be interpreted as standalone heart-failure diagnoses.}
\label{tab:staging}
\setlength{\tabcolsep}{6pt}
\fontsize{8}{9}\selectfont
\begin{tabular}{lccc}
\toprule
Model & Different stratum / $n$ & Rate & 95\% CI \\
\midrule
B1 & 27/50 & 54\% & 39--68\% \\
P2 & 22/50 & 44\% & 30--59\% \\
\bottomrule
\end{tabular}
\end{table}

\bibliographystyle{splncs04}
\bibliography{references}

\end{document}